\documentclass[10pt,a4paper]{article}
\usepackage[utf8]{inputenc}
\usepackage{amsmath}
\usepackage{amsfonts}
\usepackage{amssymb}
\usepackage{graphicx}
\usepackage{xcolor}

\usepackage{tikz}
\usepackage{enumitem}

\usepackage{amsmath}
\usepackage{amsfonts}
\usepackage{amssymb}
\usepackage{amsthm}
\numberwithin{equation}{section}

\allowdisplaybreaks

\usepackage[authoryear,round,longnamesfirst]{natbib}
\usepackage{titlesec}

\usepackage{setspace}


\usepackage[pagebackref,colorlinks=true,urlcolor=blue,
	citecolor=darkgray,linkcolor=blue,bookmarks=true]{hyperref}

\renewcommand*{\backref}[1]{}
\renewcommand*{\backrefalt}[4]{%
  \ifcase #1 %
    No citations.
  \or
    (referenced from page #2).%
  \else
    (referenced from pages #2).%
  \fi%
}

\usepackage{authblk}
\author{Yarin Gal\thanks{\href{mailto:yg279@cam.ac.uk}{yg279@cam.ac.uk}}}
\affil{University of Cambridge}
\date{September 2013}
\title{Semantics, Modelling, and the Problem of Representation of Meaning -- \\a Brief Survey of Recent Literature}

\usepackage{setspace}
\begin{document}
\maketitle

Over the past 50 years many have debated what representation should be used to capture the meaning of natural language utterances. This modelling problem forms the first step towards the extraction of information conveyed in a sentence. But even the question of what a representation of meaning should satisfy is subject to dispute.

\citet{jurafsky2008speech} for example describe a list of requirements collected from many sources.  Until recently this list has been considered rather complete -- they ask for a representation to allow us to determine the truth of a proposition and support unambiguous representation. They then ask for it to support logical inference and to be sufficiently expressive. This would allow us to reason given a sequence of declarative statements. For example, given the sentences ``Bob went to the pub'', ``It was raining outside'', and ``Bob did not have an umbrella'', we might infer that Bob got wet. However these requirements are lacking in many ways.

Following needs raised in recent research we can extend on the requirements gathered by \citet{jurafsky2008speech}.
The requirements \citet{jurafsky2008speech} describe capture no uncertainty or connotation for example.
We can demand a representation to allow us to extract the meaning of an utterance in a manageable way -- there must exist an algorithm that would allow us to parse representation from data. 
This is an important computational requirement and for scaling to large amounts of data such an algorithm has to be efficient.
Furthermore we might demand that a representation would allow us to:
\begin{itemize}[noitemsep,topsep=0pt]
\item infer the sentiment of an utterance,
\item capture uncertainty in different propositions,
\item relate utterances of similar meaning to one-another while distinguishing distinct ones, 
\item capture activities as well as agents taking part in these activities, 
\item answer questions using meaning in addition to knowledge,
\item and generate new utterances of the same meaning.
\end{itemize}
Here we will survey some of the more interesting representations proposed in recent years that try to answer some of or all these requirements. We will comment on the ways in which the different representations satisfy them. Suggestions for future research are then given based on the strengths and weaknesses of these representations. 

Throughout the survey we will give references to further reading on the various topics presented. However since the range of fields covered is rather large, the reader is assumed to be familiar with the essentials at an elementary level. These include introductory level semantics, first order logic, lambda calculus, and probability theory. In addition to that general familiarity with database query languages, probabilistic modelling, and tensor algebra is assumed. The reader should not be wary though. These topics, even though required for an in-depth understanding of each representation, are used at high level throughout the paper. We will give pointers for further reading where appropriate.

In the next section we will briefly discuss some of the traditional representations. We will extend on that with representations that develop on the compositional setting. Afterwards probabilistic extensions are discussed and then distributional approaches are mentioned briefly. The recently proposed compositional distributional representation is examined last.

\section{Traditional and Compositional representations}
First order logic is a well studied form for meaning representation. It uses 3 basic building blocks: constants (alternatively \emph{atoms}), functions, and variables, each pointing to an object in the world. Constants, such as $Cat$, can be thought of as objects in the world, while functions such as $Red(\cdot)$ act on the constants and modify them (to obtain, for example, a new atomic value $Red(Cat)$). Variables allow us to make statements and perform prediction without referring to a specific object. Using these 3 simple building blocks and the existential quantifier together with conjunction and negation, we can build complex expressions such as 
$$
\exists X (Owns(X,Red(Cat)) \wedge Feeds(X,Red(Cat),Yellow(Tuna))).
$$
interpreted as ``There exists an entity that owns a red cat and that feeds the red cat with yellow tuna''.

We can perform logical inference and entailment on such statements and using lambda expressions determine the truth value of a proposition. This representation also allows us to capture utterances in an unambiguous way; We can distinguish identical words referring to different concepts by the use of multiple atoms. For example, the word $Bank$ can refer to both the financial institution and the land alongside a river, in which case different atoms can be used to distinguish between the two: $Bank_{institution}$ compared to $Bank_{river}$. Furthermore the representation is sufficiently expressive for the modelling of many statements. 

Other representations such as Semantic Networks and frame based representations offer similar advantages to the one above. Frame based representations can be used for \emph{role labelling} in which one extracts predicates and their arguments. For example, in the sentence ``I eat what I see'' the predicate \emph{eat} would be extracted with its arguments \emph{I} and \emph{what I see}. The predicates are then classified into different groups of tasks (\emph{to eat}), and the arguments are classified into their respective roles (such as \emph{agent, theme}, and \emph{recipient}). An in-depth explanation of these two representations is given in \citep[chapter 18]{jurafsky2008speech}

The process of capturing first order logic and frame based representations from utterances is not trivial however. Different approaches have been attempted over the years, many of which fail to generalise, and all of which fail to scale. \citet{piantadosi2008bayesian} have tried to capture the first order representation of utterances limited to several dozens of words by building a probabilistic model that models possible grammars and penalises too complex ones. They then explored all possible context-free grammars capturing the utterances supplied in a recursive manner up to a certain depth in search for the best one explaining the data. This approach results in a non-scalable model for compositional semantics that can capture toy example grammars but is not suitable for practical use. 

\citet{liang2013learning} have suggested reducing the task to a simpler problem of constraint satisfaction, and by that allowed for efficient inference to be carried out. They developed a new representation called DCS trees -- dependency based compositional semantics trees -- and map utterances to these trees in an efficient way through the constraints of pre-specified trigger words and a predicate dictionary given in advance (for example the word \textit{city} would trigger the predicate \texttt{city}). Although the model was assessed on small scale data-sets with some limitations for scalability originating from the specification of trigger words, it can easily be scaled-up to map sentences over much larger domains through the use of role-labelled input and shallow extraction of predicates from text. Together with recently published large scale fact data-sets such as Freebase \citep{bollacker2008freebase} this model will allow for scalable question answering to be carried out extending on existing question answering systems that rely on simple role-labelling and empty slot completion based on previously observed statements. 

However the developed representation, studied thoroughly in \citep{liang11dcs}, lacks a formal proof for the expressive power it offers. It is unknown yet whether all sentences expressible in first order logic can be captured using DCS trees and their possible extensions. Furthermore, the formulation of DCS is analogous to and inspired by that of database querying languages (from personal communication with Liang). But even for these well studied database querying languages it has been shown that the expressiveness offered is very limited \citep{LibkinDatabaseTheory2001}. It has been shown that small differences in the language definition affect the behaviour of the language dramatically and that these languages cannot define recursive queries regardless of the aggregate functions and arithmetic operations specified (these recursive structures can, however, be observed in natural language). Furthermore, it was shown that the problem of proving expressiveness bounds for these is as hard of a problem as some long-standing open problems in complexity theory \citep{LibkinDatabaseTheory2001}. 

The use of a database querying languages for the representation of natural language utterances and not just knowledge by itself is a very interesting idea that allows for efficient logical inference and entailment to be carried out in the form of constraint satisfaction, and indeed has been studied before \citep{GiordaniNatural2010,monteleonenatural,giordani2009syntactic,giordani2010corpora}. However, even when justified with the explicit assumption placed in advance of capturing only a subset of language utterances, there is still the question of which representation to choose. The mapping of utterances to the SQL query language offers the advantages of the well studied language\footnote{For further reading on the SQL query language see \citep{date1997guide}}. The expressive power of DCS is still an open problem to be explored. 


All the representations above can be thought-of as \emph{rigid} representations as they do not capture uncertainty in the logical inference and entailment process. 
For this reason \citet{poon2009unsupervised} have suggested a \emph{probabilistic} representation to do so. The representation used was chosen to be \emph{Markov logic} \citep{domingos20071}, a representation that allows one to capture logical entailment as well as uncertainty. However, since the representation was limited to its most basic form allowing no complex first order logic statements to be captured, follow-up research by \citet{titov2011bayesian} has neglected it in favour of a simple role-label representation, where hierarchical Pitman-Yor processes were used to model statistical dependencies between predicates and their arguments\footnote{For further reading on probabilistic modelling see \citep{murphy2012machine,bishop2006pattern}}. 
There are many other alternative representations suggested for probabilistic logic; some of them will be discussed next.

\section{Probabilistic Logic representations}
The extension of compositional representations into the probabilistic domain allows us to capture uncertainty in the form of epistemic probabilities, but also raises the problem of selecting the appropriate probabilistic logic representation among the many different representations proposed over the years: Markov Logic Networks (MLN), Inductive Logic Programs (ILP), Probabilistic Logic Programs (PLP), Bayesian Logic Programs (BLP), Probabilistic Context Free Grammars (PCFG), Stochastic Logic Programs (SLP), PRISM, Probabilistic Relational Models (PRM), Stochastic Relational Models (SRM), Probabilistic Similarity Logic (PSL), and many, many more. We will give a very short survey of some of the different approaches; a much more comprehensive one is given in \citep{getoor2007introduction,de2003probabilistic}.

In probabilistic logic there exist two contrasting interpretations that can be used \citep{de2003probabilistic}. In the first we define probabilities over different worlds. Each atom in the world can take different values with different probabilities, and each assignment satisfying all atoms induces a world with a certain probability. 
For example, if the possible atoms include only $A$, $B$, and $C$ that take values from the sets $\{a_1, a_2\}$, $\{b_1, b_2\}$, and $\{c_1, c_2\}$ respectively each with probability $0.8$ for the first value and $0.2$ for the second, the probability of the world in which $A=a_1$, $B=b_1$, and $C=c_1$ would be $0.8*0.8*0.8=0.512$.
In the second interpretation, we assign probabilities to different derivations. Atoms are fixed, but the entailments are uncertain. For example, we might say that given atom $U$, the atom $V$ might follow with probability $0.7$, or the atom $W$ might follow with probability $0.3$. This induces probabilities over proofs, in which different derivations have different probabilities. Probabilistic Context Free Grammars (PCFG) and Stochastic Logic Programs (SLP, \citet{muggleton1996stochastic}) belong to this interpretation, while most of the other representations belong to the former interpretation.

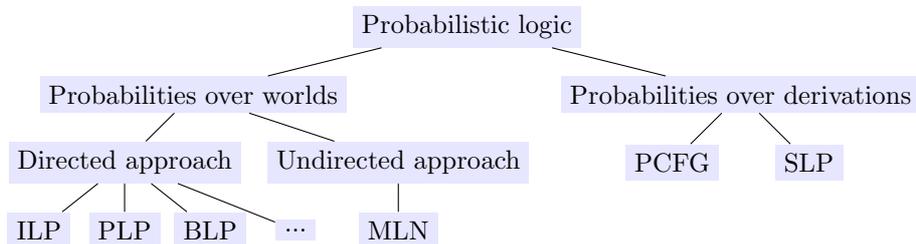
\begin{figure}
\begin{tikzpicture}
  [scale=0.9,auto=center,every node/.style={rectangle,fill=blue!10}]
  \node (n1) at (5,3) {Probabilistic logic};
  \node (n2) at (1,2) {Probabilities over worlds};
  \node (n6) at (0,1) {Directed approach};
  \node (n8a) at (-1.25,0) {ILP};
  \node (n8b) at (0,0) {PLP};
  \node (n8c) at (1.25,0) {BLP};
  \node (n8d) at (2.5,0) {...};
  \node (n7) at (4,1) {Undirected approach};
  \node (n9) at (4,0) {MLN};
  \node (n3) at (9,2) {Probabilities over derivations};
  \node (n4) at (8,1) {PCFG};
  \node (n5) at (10,1) {SLP};

  \draw (n1) -- (n2);
  \draw (n1) -- (n3);
  \draw (n3) -- (n4);
  \draw (n3) -- (n5);
  \draw (n2) -- (n6);
  \draw (n2) -- (n7);
  \draw (n6) -- (n8a);
  \draw (n6) -- (n8b);
  \draw (n6) -- (n8c);
  \draw (n6) -- (n8d);
  \draw (n7) -- (n9);
\end{tikzpicture}
\caption{Relations between the different representations in Probabilistic Logic}
\label{fig:rel-prob-log}
\end{figure}

In the former interpretation there exist two main approaches that model the probabilities over clauses differently \citep{cussens2007logic}. The directed approach assumes that there exists a kernel set of logical clauses all of which have probabilities explicitly defined for them. The rest of the clauses have their probabilities induced recursively. 
For example, if we know that $C=c_1$ if either $A=a_1$ or $B=b_1$ in the example above, then the probability of the world in which $A=a_1$, $B=b_1$, and $C=c_1$ would be in this case $0.8*0.8=0.64$ since $C$ is determined by $A$ and $B$, and the probability of $C$ taking the value $c_1$ would be $1-0.2*0.2=0.96$.
Most of the representations above use this approach. In the undirected approach no clause has its probability explicitly stated. In this approach the probability of each possible world is given by a set of its `features' where each feature has a real valued parameter associated with it. Markov Logic Networks \citep{domingos20071} adhere to this approach, where the features are the number of true ground instances of a clause in a given world. These relations between the different representations are summarised in figure \ref{fig:rel-prob-log}.

Probabilistic Logic Programs and their extension to Bayesian Logic Programs \citep{kersting20071} are examples of directed models. Bayesian Logic Programs are composed of a logical component which is a set of clauses of a specific form induced by a Bayesian directed graph, and a quantitative component which describes the conditional probability distributions in the graph and combining rules. We can represent these as a probabilistic Prolog program (a set of definitive logical clauses). \citet{puech2003comparison} have studied the expressive power of these representations and compared them to the expressive power of SLPs (which define probabilities over derivations) -- showing that SLPs can encode the same knowledge as a sub-class of BLPs.  Inference in BLPs is much easier than in SLPs as well, using structured EM to find the structure of the BLP; inference in SLPs is known to be a very hard problem \citep{de2003probabilistic}.

The proposed Markov logic networks representation was an attempt by \citet{domingos20071} to unify the different approaches developed in the field of statistical relational modelling by suggesting a framework in which first order logic clauses hold weights and thus represent a probability distribution over possible worlds. Leading figures in the field have adopted this approach \citep{huynh2008discriminative,mihalkova2007mapping,mihalkova2007bottom}, however due to the difficulties in inference this approach is losing its popularity among its advocates \citep{beltagy2013montague}. 

The representations above can capture epistemic uncertainty in the logic, an advantage over traditional compositional representations; however these are \emph{discrete}, and offer no way to reason about \emph{similarity} between sentences. The recently suggested Probabilistic Similarity Logic \citep{brocheler2012probabilistic} tries to accommodate for that by introducing a metric for distances between the atoms. Other approaches have adopted a different route, where a \emph{continuous} representation is used to capture such atoms.

\section{Distributional representation}
Distributional methods have been in use in \emph{lexical} semantics for more than 40 years \citep{jurafsky2008speech}. They make use of a continuous representation for the words; For each word its neighbouring words are collected resulting in long binary vectors, often having their dimensionality reduced. Different metrics can be used with such \emph{co-occurrence} vectors, where the $l_2$ metric is a common choice to capture the distance between words. This can then be used for word-sense disambiguation, hyponymy, and other word relation tasks. We cannot capture complex structures with this representation though, as only the individual words of a sentence are represented as vectors of real numbers. To capture more complex behaviour, we have to turn to a \emph{compositional} distributional representation.

\section{Compositional Distributional representation}
\citet{coecke2010mathematical} have recently suggested a novel representation relying on a combination of the traditional compositional representation and the continuous distributional one. Their work was influenced by \citet{baroni2010nouns}'s work which suggested the representation of nouns as vectors in a continuous space, and adjectives as matrices linearly transforming these nouns. \citet{coecke2010mathematical}'s model was originally developed as an extension of theoretical work in quantum informatics, abstracted away from the field into category theory, and cast back into the field of linguistics \citep{heunen2013quantum}, by which it was shown to correspond to some very familiar structures in the field of linguistics such as Montague grammar \citep{clark2012vector}. 

The representation uses tensor products\footnote{For an introduction to tensor algebra see \citep{lang2002algebra}} to carry information from word level to a higher level, and then uses linear functions on these to project into lower dimensional sentence spaces. For example, the nouns ``Dogs'' and ``Cats'' might be represented as vectors $d$ and $c$, and the verb ``chase'' might be represented as the third-order tensor $T$. The sentence ``Dogs chase cats'' can then be represented as the tensor product $d \otimes T \otimes c$, where $\otimes$ is the Kronecker product. We exploit the compositionality of this product to obtain a vector in the sentence space. We do this by reducing the dimensionality of the product of the 2 vectors and the third-order tensor with a linear projection. \citet{grefenstette2010concrete} examined the problem of finding concrete tensor and vector values for such representation for a given data set by looking at the third-order tensor as ``taking'' a noun from the left, ``taking'' a noun from the right, and looking at the properties of such nous from the distributional representation. For example, if the second dimension of the noun vector corresponds to the property ``likes chasing small fluffy animals'', the third and fourth dimensions correspond to the properties ``is small'' and ``is fluffy'', and if a cat is characterised as small and fluffy in its representation, while a dog is characterised as likes to chase small and fluffy animals, then the tensor representing ``chase'' should take vectors from the left that have the ``likes chasing small fluffy animals'' property and vectors from the right that have the ``is small'' and ``is fluffy'' properties, and after reducing the dimensionality return a sentence vector encapsulating all these characteristics. In the example given in \citet{grefenstette2010concrete}, the sentence space represented the ``truthness'' value of a proposition. The sentence space is constructed as having one dimension representing how true a proposition is. The proposition ``dogs chase cats'' would thus be mapped to a high ``truthness'' value, while the proposition ``cats chase dogs'' would be mapped to a low one.

\citet{grefenstette2013towards} has shown that this representation can be used to capture uncertainty as well. He explains how the use of linear projections can be identified with non-quantified first order logic. However he also proves a negative result showing that quantified first order logic cannot be obtained when using only linear projections. He then introduces non-linear mappings to obtain quantified first order logic. 

This representation has been gaining support in the field of semantic modelling \citep{van2013tensor,hermann2013not}. It offers unambiguous representation, it is expressive, it might allow the inference of sentiment and role labels (as we will see below), captures uncertainty and the truth of a sentence, and models the distance between sentences as points in a sentence space thus captures sentence similarity. Current research is looking into logical inference and the extraction of the meaning of an utterance into the sentence space in a manageable way. 

Future research into generative models for this representation will allow us to generate sentences from points in a sentence space as well as capture the meaning of an utterance into the sentence space, and research into non-linear reductions of the tensor product will allow logical inference to be carried out, and as such the creation of question-answering systems that go beyond role-label processing and constraint satisfaction.

\section{Conclusions}

There are still many open problems one needs to solve to use the promising Compositional Distributional representation. Other developments of compositional representations that make use of distributional elements have been studied empirically though, and managed to obtain state-of-the-art results in many tasks in natural language processing, including sentiment analysis on adverb-adjective pairs and movie reviews \citep{socher2011semi,socher2012semantic}, role labelling on semantic relationships \citep{socher2012semantic}, paraphrase detection on the MSRP paraphrase corpus \citep{socher2011dynamic}, syntactic parsing on the Penn Treebank \citep{socher2013Parsing}, and word similarity tasks \citep{luong2013better}. 

This suggests that a theoretically grounded development of the field would be of great use. Light could be shed on the success of the empirical research, while suggesting new extensions to pursue further.

\section*{Acknowledgements}

The author would like to thank Richard Socher, Professor Percy Liang, Doctor Edward Grefenstette, and Professor David McAllester for some interesting discussions regarding the different topics brought above.

\newpage

\bibliographystyle{plainnat}
\bibliography{../ref}

\end{document}